\documentclass[review]{elsarticle}

\usepackage{lineno,hyperref}
\usepackage{array}
\usepackage{amssymb}
\usepackage{enumerate}
\usepackage{makecell}
\usepackage{mathtools}
\usepackage{array}
\usepackage{multirow}
\usepackage{ulem}
\usepackage{color}
\usepackage{diagbox}
\usepackage{enumerate}
\usepackage{graphicx}
\usepackage{subfigure}
 \usepackage{tikz,mathpazo} 
\usepackage{indentfirst}
\usepackage{soul}
\usepackage{amsmath}
\usepackage[justification=centering]{caption}
\usepackage{tabularx}
\usepackage{algorithm}
\usepackage{algorithmicx}
\usepackage{algpseudocode}
\usepackage{adjustbox}
\usepackage{geometry}
\usetikzlibrary{shapes.geometric, arrows}
\modulolinenumbers[5]
\newtheorem{definition}{Definition}[section]
\newtheorem{example}{Example}[section]

\soulregister{\cite}7
\soulregister{\ref}7

\newcommand{\circledsmall}[1]{\lower.7ex\hbox{\tikz\draw (0pt, 0pt)%
    circle (.5em) node {\makebox[0.1em][c]{\small#1}};}}

\newcommand{\circledtiny}[1]{\lower.7ex\hbox{\tikz\draw (0pt, 0pt)%
    circle (.3em) node {\makebox[0.1em][c]{\tiny #1}};}}

\journal{Journal of \LaTeX\ Templates}

\begin{document}

\begin{frontmatter}

\title{Belief Evolution Network-based Probability Transformation and Fusion}

    \author[a]{Qianli Zhou}
    \ead{zhou\_qianli@hotmail.com}
    \author[a]{Yusheng Huang}
    \ead{huangyusheng@std.uestc.edu.cn}
    
    \author[a,b,c,d]{Yong Deng\corref{mycorrespondingauthor}}
    \cortext[mycorrespondingauthor]{Corresponding author}
    \ead{dengentropy@uestc.edu.cn}
    \address[a]{Institute of Fundamental and Frontier Science, University of Electronic Science and Technology of China, Chengdu 610054, China}
    \address[b]{School of Eduction Shannxi Normal University, Xi’an, 710062, China}
    \address[c]{School of Knowledge Science, Japan Advanced Institute of Science and Technology, Nomi, Ishikawa 923-1211, Japan}
    \address[d]{Department of Management, Technology, and Economics, ETH Zurich, Zurich, 8093, Switzerland}

\begin{abstract}
Smets proposes the Pignistic Probability Transformation (PPT) as the decision layer in the Transferable Belief Model (TBM), which argues when there is no more information, we have to make a decision using a Probability Mass Function (PMF). In this paper, the Belief Evolution Network (BEN) and the full causality function are proposed by introducing causality in Hierarchical Hypothesis Space (HHS). Based on BEN, we interpret the PPT from an information fusion view and propose a new Probability Transformation (PT) method called Full Causality Probability Transformation (FCPT), which has better performance under Bi-Criteria evaluation. Besides, we heuristically propose a new probability fusion method based on FCPT. Compared with Dempster Rule of Combination (DRC), the proposed method has more reasonable result when fusing same evidence.
\end{abstract}


\end{frontmatter}

\section{Introduction}
\label{intro}
As a generalization of the Probability Theory (ProbT), Dempster-Shafer Theory (DST) is proposed by Dempster \cite{dempster2008upper} based on the mapping of upper and lower probabilities, and then developed by Shafer \cite{shafer1976mathematical} into a mathematical theory to handle uncertain information. By assigning belief on the power set of the Frame of Discernment (FoD), DST has a stronger ability to model underlying information than ProbT. In terms of information handling, Dempster Rule of Combination (DRC) can not only be backward compatible with ProbT, but have Matthew effect as well. In discrete-event systems, DRC and its extensions are core algorithms in the field of information fusion \cite{Xiong2021InformationSciences,bronevich2021measures}, which are widely used in Evidential Reasoning Approaches \cite{yang2001rule,yang2002evidential,yang2006belief}, expert decision making \cite{zhou2020weight,liu2022multiattribute}, reliability analysis \cite{wang2019risk,gao2021NET,Chenxingyuan2022SREM} and medical diagnosis \cite{cheong2013construction,Xiao2022Generalizeddivergence,tao2021dynamic},. Basic Probability Assignment (BPA), as a basic information unit of DST, allows belief on the multi-element subsets. Smets \cite{smets1994transferable} proposes the Transferable Belief Model (TBM) to allow the belief on empty set not be zero, and interpret the focal element with multiple elements as ignorance, i.e., the belief can not be assigned more accurate. Based on the above, the uncertain machine learning methods are developed under the framework of DST. Liu \textit{et al.} \cite{liu2021new,9057440,9108588} optimize the evidential classifier under different FoD; Ma \textit{et al.} \cite{ma2016online} propose the evidential decision tree based on the likelihood belief functions; Den{\oe}ux \textit{et al.} \cite{denoeux2016evidential,gong2021evidential} propose the evidential clustering methods based on the fuzzy c means. To represent uncertain information more comprehensively, DST-based extensions have also been a hot research topic in recent years. Xiao \cite{Xiao2022NQMF} extends DST to complex number space to predict interference effects \cite{Xiao2021CEQD}. Bronevich \cite{Bronevich2018modeling} proposes the Generalized Credal Set based on the imprecise probability interpretation of DST. Deng \cite{deng2022random} replaces the combination space (power set) with the permutation space and proposes the Random Permutation Set to represent the ordered belief information.

Since the elements within FoD are mutually exclusive, Probability Transformation (PT) is the necessary step of decision making when no more information is available. How to transform BPA to Probability Mass Function (PMF) has been researched from the different perspectives. As the decision layer in TBM, the Pignistic Probability Transformation (PPT) is proposed by Smets \cite{smets2005decision}. Cobb and Shenoy \cite{cobb2006plausibility} utilize the normalized plausibility function of elements to propose the Plausibility Transformation Method (PTM), which is the only PT that satisfies the consistency of CRD. Sudano \cite{martin2006yet} redistributes the belief of multi-element focal elements to generate the probability distribution (PraPl). Dezert and Smarandache \cite{dezert2008new} propose a PT which can adjust the degree of optimism based on Dezert-Smarandache Theory (DSmP). Cuzzling is committed to study the visual DSET \cite{cuzzolin2021geometry}, and proposes probability transformation from the perspective of graphs (CuzzP) \cite{cuzzolin2012relative}. Facing these methods, Dezert \textit{et al.} \cite{dezert2012hierarchical} propose $3$ requirements of PT and Han \textit{et al.} \cite{han2015evaluation} extend them to establish a Bi-Criteria evaluation method to fit different applications. According to Bi-Criteria evaluation method, the result of PT should be both similar to the original BPA and should have less Probability Information Content (PIC) to be more beneficial for decision making. Jiang \cite{jiang2019new} has demonstrated that under the criteria of belief correlation coefficient \cite{jiang2018correlation}, PMF generated by PPT is the most similar result with origin BPA. However, the PPT is evenly assigned the belief of focal elements, resulting in its higher PIC. In DSMP, the most optimistic transformation leads to the least PIC, however the similarity with the original BPA is low. Therefore, there is no PT method to satisfy both of requirements.

Considered from the perspective of total uncertainty \cite{Qiang2022fractal}, BPA transformed to PMF is a process of uncertainty reduction. Therefore, PT can also be regarded as a process of information fusion. The most commonly used combination rule in DST is the Dempster Combination Rule (DRC). For independent and credible Bodies of Evidence (BoEs), DRC is the most appropriate choice. However, for conflicting BoEs, Zadeh gives the convergence paradox of DRC. To resolve the paradox, Yager \cite{yager1987dempster} assigns the conflicts in the fusion process into the vacuous set. For unreliable BoEs that cannot be quantified, Dubois and Prade \cite{dubois2008set} propose the Disjunctive Combination Rule (DCR) based on set-theoretic. Dezert and Smarandache \cite{smarandache2006advances} propose DSMT to resolve the fusion of conflicting BoEs in higher dimensions, however, it usually entails high computational complexity. Murphy averages the the BoEs to mitigate the impact of conflicts, and continuously fuses the modified BoEs. Den{\oe}ux \cite{denoeux2008conjunctive} proposes Cautious Combination Rules to deal with BoEs generated from non-independent sources.In decision-making, Yang and Xu \cite{yang2013evidential} introduce power sets in the FoD and utilize Weight $w$ and Reliability $r$ to propose Evidential Combination Rule (ECR). The existing combination rules are based on the reliability and importance of the source of evidence to resolve conflicts between BoEs. In essence, it does not break through the intersection idea of DRC. The DRC-based information fusion algorithms also have the characteristic of the Matthew effect. When combining the same information, the information will gradually tend to polarize until the belief of a certain target is $1$. So even for a target with a small support, as long as its support is greater than other targets, after a finite number of fusions, its belief will reach $1$.

The previous methods of PT and information fusion both operate on the entire power set of FoD. Gordon and Shortliffe \cite{gordon1985method} propose the method to handle BoEs on Hierarchical Hypothesis Space (HHS). The Hierarchical processing approach is similar to the granular computing \cite{bargiela2016granular}. It has achieved better application in multi-granularity decision problems. In this paper, we introduce the causality in HHS and propose a directed acyclic network to represent BPA called Belief Evolution Network (BEN). According to the BEN, we propose the full causality ($FC$) function to express uncertainty of BPA from a new perspective. We interpret the PT as a information fusion process on BEN, and propose a new PT method called Full Causality Probability Transformation (FCPT). Under the Bi-Criteria, the FCPT is superior to existing PT methods. Finally, we combine the new transformation method and DCR to propose a new probability fusion method called FCPT-based Combination Rule (FPCT-CR). 

The rest of paper is organized as follows: Section \ref{pre} introduces the basic concepts of DST and the common combination rules and PT methods. Section \ref{ben} proposes the BEN and defines a new belief function called $FC$ function. In Section \ref{benpts}, we interpret PPT on BEN and propose the FCPT. A new probability fusion method is proposed in Section \ref{DTCR}, which has better performance than DRC in classification. In Section \ref{con}, we summarize the whole paper and discuss the future research directions of BEN.

\section{Preliminaries}
\label{pre}

\subsection{Dempster-Shafer Theory}
For a finite set $\Theta$ called Frame of Discernment (FoD) with $n$ mutually exclusive elements, DST allows assigning belief on its power set $2^{\Theta}=\{\{\emptyset\},\{\theta_1\},\dots,\{\theta_n\},\{\theta_1\theta_2\},\dots,\{\theta_1\dots\theta_n\}\}$.
\subsubsection{Basic Probability Assignment}

\begin{definition}[BPA]\label{BBA}\cite{dempster2008upper,shafer1976mathematical}
\rm{For an $n$-element FoD $\Theta$, its basic information unit in DST is Basic Probability Assignment, also called mass function $m$. For a normal mass function, $m$ satisfies
\begin{equation}
\sum_{F \in 2^\Theta\setminus\emptyset }m(F)=1;~m(F)\in[0,1];~m(\emptyset)=0.
\end{equation}
$m(F)$ represents the agents' support to subset $F$, and there are no more information to assign it to subsets, and when $m(F)\neq 0$, $F$ is the focal element. }
\end{definition}

In order to convenient to expression, there are some special BPAs.
\begin{itemize}
    \item When there is only one focal element $\Theta$, i.e., $m(\Theta)=1$, $m$ is \textbf{vacuous mass function}, which means total ignorance in TBM.
    \item When focal elements are singletons, $m$ is \textbf{Bayesian mass function}, which can be seen as a PMF.
    \item When focal elements satisfy $G\subset F$ or $F \subset G$, $m$ is \textbf{consonant mass function}, which can be transformed to the possibility distribution uniquely.
\end{itemize}

\subsubsection{Belief functions}

Different from ProbT, in addition to BPA, belief functions also are basic information units in DST.

\begin{definition}[belief functions]
\rm{For a BPA $m$ under an $n$-element FoD $\Theta$, its belief ($Bel$) function, plausibility ($Pl$) function and commonality ($Q$) function are
\begin{equation}\label{bel_eq}
\begin{aligned}
& Bel(F)=\sum_{\emptyset\neq G\subset F}m(F)=1-m(\emptyset)-Pl(F^c);\\
 &   Pl(F)=\sum_{G \cap F \neq \emptyset}m(F)=1-m(\emptyset)-Bel(F^c);\\
  &  Q(F)=\sum_{F\subset G}m(G);\\
   & m(F)=\sum_{F\subset G}(-1)^{|G|-|F|}Q(G)=\sum_{G \subset F}(-1)^{|F|-|G|}b(G);
\end{aligned}
\end{equation}
where $F^c$ is the complement of $F$ under $\Theta$, and $b(F)=m(\emptyset)+Bel(F)$.
}
\end{definition}
 According to Eq.(\ref{bel_eq}), there are $2$ couples of dual functions. For non-empty focal elements, $Bel$ and $Pl$ functions are dual, where $Bel(F)$ means the support to all elements $\theta\in F$, and $Pl(F)$ means the non-negative degree of $F$, which has same meaning with 'possibility'. They compose the belief interval $BI(F)=[Bel(F),Pl(F)]$, which can be seen as the probability domain of credal set. In addition, $b$ and $Q$ functions are dual and satisfy $b^c(F)=Q(F^c)$.

\subsubsection{Evidence combination rules}

For BoEs generated from independent sources, if sources are credible, Conjunctive Combination Rule (CCR) and Dempster Rule of Combination (DRC) are used to fuse them.

\begin{definition}[CCR\&DRC]\label{ccr}\cite{denoeux2008conjunctive}\rm{
For an $n$-element FoD $\Theta$ with BPAs $m_1$ and $m_2$ from independent agents (sensors or attributes). Conjunctive Combination Rule (CCR) $m_1 \circledsmall{$\cap$} m_2$ is defined as
\begin{equation}\label{ccre1}
m_{1\circledtiny{$\cap$}2}(F)=\sum_{G\cap H=F}m_1(G)m_2(H).
\end{equation}
Using $Q$ function can simplify the calculation process
\begin{equation}
    Q_{1\circledtiny{$\cap$}2}(F)=Q_1(F)\cdot Q_2(F).
\end{equation}
Since the belief on the empty set lacks real physical meaning, the DRC normalizes the results of the CCR.
\begin{equation}
  m_{1\oplus 2}(F_i)=
\begin{cases}
K^{-1}\cdot m_{1 \circledtiny{$\cap$} 2}(F)& \text{$F_i$ $\neq$ $\emptyset$}\\
0& \text{$F$ = $\emptyset$}
\end{cases},  
\end{equation}
where $K=\sum_{G \cap H =\emptyset}m(G)m(H)$ is conflict coefficient. 
}
\end{definition}

When the independent sources are not credible and their reliability can be quantified. Evidential Combination Rule (ECR) can deal with them reasonably.

\begin{definition}[ECR]\cite{yang2013evidential}
\rm{
For an $n$-element FoD $\Theta$, two BPAs $m_1$ and $m_2$ with reliability $r_i$ and weight $w_i$ ($i=1,2$), their combination results $m$ of ECR is
\begin{equation}
   m(F)=[(1-r_2)\widetilde{m_1}(F)+(1-r_1)\widetilde{m_2}(F)]+\sum_{G\cap H=F}\widetilde{m_1}(G)\widetilde{m_2}(H), 
\end{equation}
where $\widetilde{m_i}$ is
\begin{equation}
    \widetilde{m_i}(F)=
\begin{cases}
0 &F=\emptyset\\
c_{rw,i}\cdot m(F)& F\in 2^\Theta\setminus\{\emptyset\}\\
c_{rw,i}(1-r_i) & F=2^\Theta
\end{cases}, 
\end{equation}
where $c_{rw,i}=1/(1+w_i-r_i)$. When the $r_i$ and $w_i$ does not exist, the ECR degenerates to DRC.
}
\end{definition}

When the reliability of sources can not be quantified, and we only know that there are at least one source is credible, Disjunctive Combination Rule (DCR) dose not deny any support information.

\begin{definition}[DCR]\label{dcr}\cite{dubois2008set}
\rm{For an $n$-element FoD $\Theta$ with BPAs $m_1$ and $m_2$ from independent agents (sensors or attributes). Disjunctive Combination Rule (DCR) $m_1 \circledsmall{$\cup$} m_2$ is defined as
\begin{equation}
m_{1\circledtiny{$\cup$}2}(F)=\sum_{G\cup H=F}m_1(G)m_2(H).
\end{equation}
Using $b$ function can simplify the calculation process
\begin{equation}
    b_{1\circledtiny{$\cup$}2}(F)=b_1(F)\cdot b_2(F).
\end{equation}
}
\end{definition}

Different from CCR, DCR is a process of entropy increasing, and its direction of belief evolution is 'ignorance'. In addition, for dependent sources, Cautious Combination Rules should be used. Because it dose not discuss in this paper, the specific definition can refer \cite{denoeux2008conjunctive}.

\subsection{Probability Transformation}

\subsubsection{Common Probability Transformation Method}

Based on different application scenarios, the method of PT varies. In this part, we introduce several general classic PT methods.

\begin{definition}[PT]\label{pt}
\rm{For an $n$-element FoD with a BPA $m$, the PT methods of $m$ is shown in Table \ref{s2t1-PT}.

\begin{table}
\begin{center}
\begin{tabular}{c|c}
  \Xhline{1.4pt}
  Methods&Expression\\
  \hline
PPT\cite{smets2005decision}&$BetP(\theta)=\sum_{\theta \in F} m(F)/(|F|\cdot(1-m(\emptyset)))$\\
PMT\cite{cobb2006plausibility}&$PnPl(\theta_i)=Pl(\theta_i)/\sum_{\theta\in \Theta}Pl(\theta)$\\
PraPl\cite{martin2006yet}&$PraPl(\theta)=Bel(\theta)+Pl(\theta)\cdot (1-\sum_{\theta\in \Theta}Bel(\theta))/\sum_{\theta\in \Theta}Pl(\theta)$\\
DSmP\cite{dezert2008new}&$DSmP_{\epsilon}(\theta)=m(\theta)+(m(\theta)+\epsilon) \cdot \sum_{F\subseteq\Theta,|F|\geq 2,\theta\in F}\frac{m(F)}{\sum_{G\subset F,|G|=1}m(G)+|F|\cdot\epsilon}$\\
CuzzP\cite{cuzzolin2012relative}&$CuzzP(\theta)=m(\theta)+\sum_{F\subseteq \Theta, |F|>1}m(F)\cdot \frac{Pl(\theta)-m(\theta)}{\sum_{\theta\in\Theta}Pl(\theta)-m(\theta)}$\\
ITP\cite{deng2020novel}&$ITP(\theta)=m(\theta)+\sum_{\theta \in F, F \subseteq \Theta}m(F) \cdot \varepsilon_{\theta}$\\
  \Xhline{1.4pt}
\end{tabular}
\end{center}
\caption{Some general classical PT methods. $\varepsilon_{\theta}$ indicates the normalization support degree of $\theta_i$ in $F_i$, specific calculation is shown in \cite{deng2020novel}.}
\label{s2t1-PT}
\end{table}}
\end{definition}

According to Table \ref{s2t1-PT}. Except PTM, the expressions of PT methods can be written as $\mathbb{P}(\theta)=m(\theta)+\sum_{{|F|>1,\theta\in F}}\beta\cdot m(F)$. Thus, for the previous PT methods, their purpose can be summarized as finding a reasonable method to assign the belief in multi-element focal elements to singletons.

\subsubsection{Evaluation of Probability Transformations}

According to the relationship of PMF and BPA in TBM. Dezert \textit{et al.} propose $3$ requirements of PT methods \cite{dezert2012hierarchical}.

\begin{enumerate}
\item \textbf{p Consistency:} When BPA degenerates to PMF, the transformed PMF is one of potential results.
\item \textbf{ULB Consistency:} The transformed PMF  $\mathbb{P}(\theta)$is supposed to fall in belief interval $[Bel, Pl]$, i.e., $Bel(\theta)\leqslant \mathbb{P}(\theta)\leqslant Pl(\theta)$.
\item \textbf{Combination Consistency:} For two BPAs $m_1$ and $m_2$, the order of DRC and PT dose not change its result. i.e., $\mathbb{P}_{m_1}\oplus \mathbb{P}_{m_2}=\mathbb{P}_{m_1\oplus m_2}$.
\end{enumerate}

According to Table \ref{s2t1-PT}, no PT method can meet all three requirements. For \textbf{combination consistency}, only PTM can satisfy, however it does not satisfy \textbf{ULB consistency}. All other methods satisfy only \textbf{p consistency} and \textbf{ULB consistency}. Therefore, these three requirements do not assess the goodness of PT methods.
In addition, Sudano \cite{sudano2015pignistic} proposes the Probabilistic Information Content (PIC) to evaluate the PTT. For a probability distribution $P=\{p(\theta_1),\dots,p(\theta_n)\}$, PIC is defined as
\begin{equation}\label{pice}
PIC(P)=1-E_\mathrm{N}=1+\frac{\sum_{i=1}^n p(\theta_i)\log p(\theta_i)}{\log n},
\end{equation}
where $E_\mathrm{N}$ is the normalized Shannon entropy. In evaluation process, the larger PIC, the more conducive to the decision-making for transformation result. Han \textit{et al.} \cite{han2015evaluation} indicates that only entropy-based criteria is not sufficient to evaluate. Similarity between original BPA and transformed PMF should also be an evaluation index. The more similarity ensures the PMF retains more information about the BPA. Based on the above, A Bi-Criteria evaluation method comprehensive the entropy and similarity is proposed. Suppose $n$ transformed PMFs $\mathbb{P}=\{P_1,\cdots,P_n\}$, the Bi-Criteria evaluation method can be written as
\begin{equation}\label{hane}
C_\textrm{joint}(\mathbb{P}_i)=\alpha \cdot E'_\mathrm{N}(\mathbb{P}_i)+(1-\alpha)\cdot d'(\mathbb{P}_i),
\end{equation}
where the $E_\mathrm{N}'(\mathbb{P}_i)$ and $d'(\mathbb{P}_i)$ are entropy index and similarity index respectively.
\begin{equation}\label{ente}
E_\mathrm{N}'(\mathbb{P}_i)=\frac{E_\mathrm{N}(\mathbb{P}_i)-min(E_\mathrm{N})}{max(E_\mathrm{N})-min(E_\mathrm{N})},
\end{equation}

\begin{equation}\label{de}
d'(\mathbb{P}_i)=\frac{d(\mathbb{P}_i)-min(d)}{max(d)-min(d)},
\end{equation}

where the $E_\mathrm{N}(\mathbb{P}_i)$ is the Normalized Shannon entropy of $\mathbb{P}_i$, and $d(\mathbb{P}_i)$ is distance of original BPA and transformed PMF \cite{jousselme2001new}. 

In addition to using distance to express the relationship between BBA and probability of transformation, the correlation coefficient proposed by Jiang \cite{jiang2018correlation} can also measure the similarity. For two BPAs $m_1$ and $m_2$ under FoD $\Theta$, the correlation coefficient is defined as
\begin{equation}\label{cc1}
r_{BPA}(m_1,m_2)=\frac{c(m_1,m_2)}{\sqrt{c(m_1,m_1)\cdot c(m_2,m_2)}},
\end{equation}
where $c(m_1,m_2)$ is 
\begin{equation}\label{cc2}
c(m_1,m_2)=\sum_{F\subseteq \Theta }\sum_{G\subseteq \Theta }m_1(F)m_2(G) \frac{|F\cap G|}{|F\cup G|}.
\end{equation}

According to Eq. (\ref{cc2}), the correlation coefficient and distance in DST both use $\frac{|F\cap G|}{|F\cup G|}$ to handle the relationship of focal elements. Hence, the correlation coefficient also can be the similarity index in Bi-Criteria evaluation.

\section{Belief Evolution Network and full causality function}
\label{ben}

In this Section, we model BPA in a directed acyclic network called Belief Evolution Network (BEN), and use causality to describe the relationship of subsets of FoD, which provides a new idea for the expression of BPA. In addition, according to BEN, we propose a new belief function, called full causality ($FC$) function, which can be seen as the full set of $b$ function and $Q$ function.

\subsection{Belief Evolution Network}

\subsubsection{Relationship of BBA and probability distribution}

\begin{figure}[htbp!]
\centering
\includegraphics[width=0.45\textwidth]{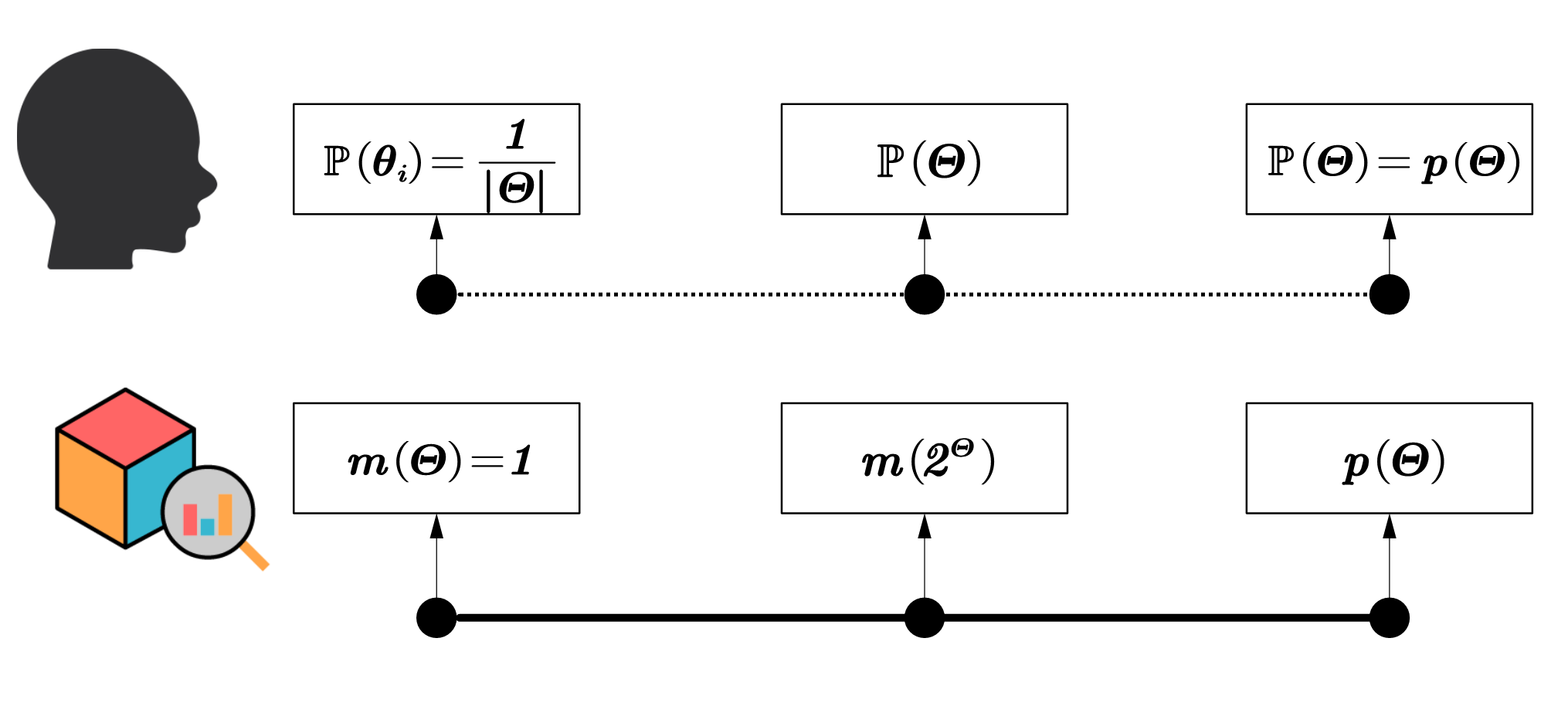}
\caption{The relationship between BBA and probability distribution at same time.}
\label{BBA_p_m}
\end{figure}

In \cite{zhou2022higher}, we use a new model to describe the relationship of BPA and PMF. For a event described by $n$-element FoD $\Theta$, a timeline can be used to describe the evolution of our understanding. The initial point of timeline represents $m(\Theta)=1$, which means that we know nothing about FoD. As time goes by, our understanding gradually becomes clear. The end point of timeline is a PMF, which means that the development of the event is over, and the distribution of elements is certain. PMF at the end point is not the same as the empirical probability, when no external information added, it dose not change with time. The points on the timeline are BPAs, which represents the development process of the event. Although the points on the timeline exist in form of BPAs, they are observed in the form of PMF, which are usually used for fusion, prediction and reasoning. For example, using the FoD $\Theta=\{\{Head\},\{Tail\}\}$ to represent the coin tossing, the probability we observe is $\{p(\{Head\})=0.5,p(\{Tail\})=0.5\}$, but because the coin tossing is still happening, its actual form should be $\{m(\{Head\})=0.5(1-\epsilon),m(\{Tail\})=0.5(1-\epsilon),m(\{Tail,Head\})=\epsilon\}$, which is consistent with Den{\oe}ux's view in \cite{denoeux2008conjunctive}. According to the above, BPA is an expression which can be convenient to represent process of information evolution, and the PMF is the form observed by people. Fig. \ref{BBA_p_m} shows their relationship at different moments.

\subsubsection{Introducing causality in Hierarchical Hypothesis Space}

\begin{figure}[htbp]
  \begin{minipage}{0.48\linewidth}
    \centering
    \includegraphics[width=0.98\textwidth]{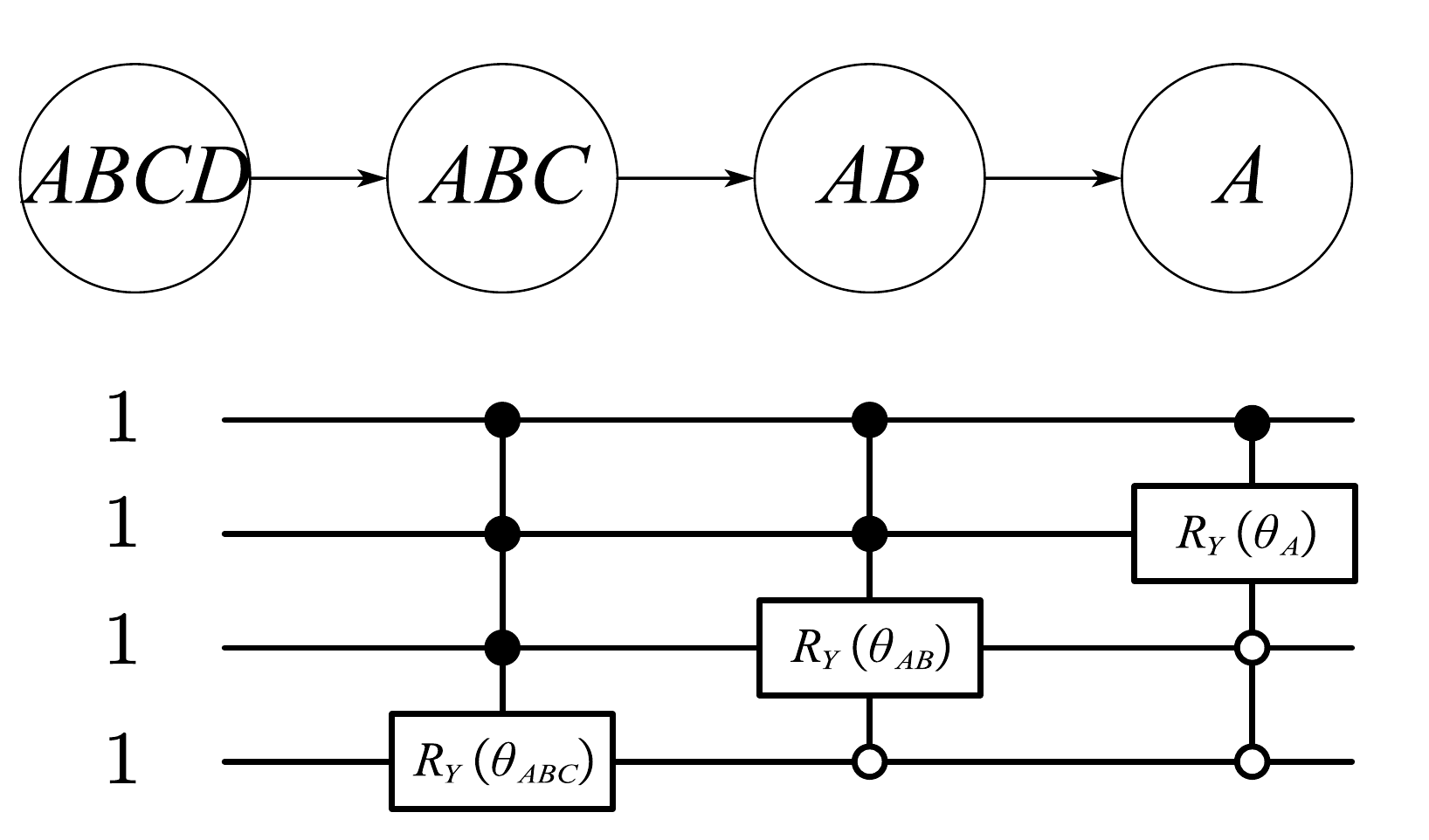}
    \caption{Quantum circuits to implement belief functions.}
    \label{Bool_3}
  \end{minipage}
    \begin{minipage}{0.48\linewidth}
    \centering
    \includegraphics[width=0.98\textwidth]{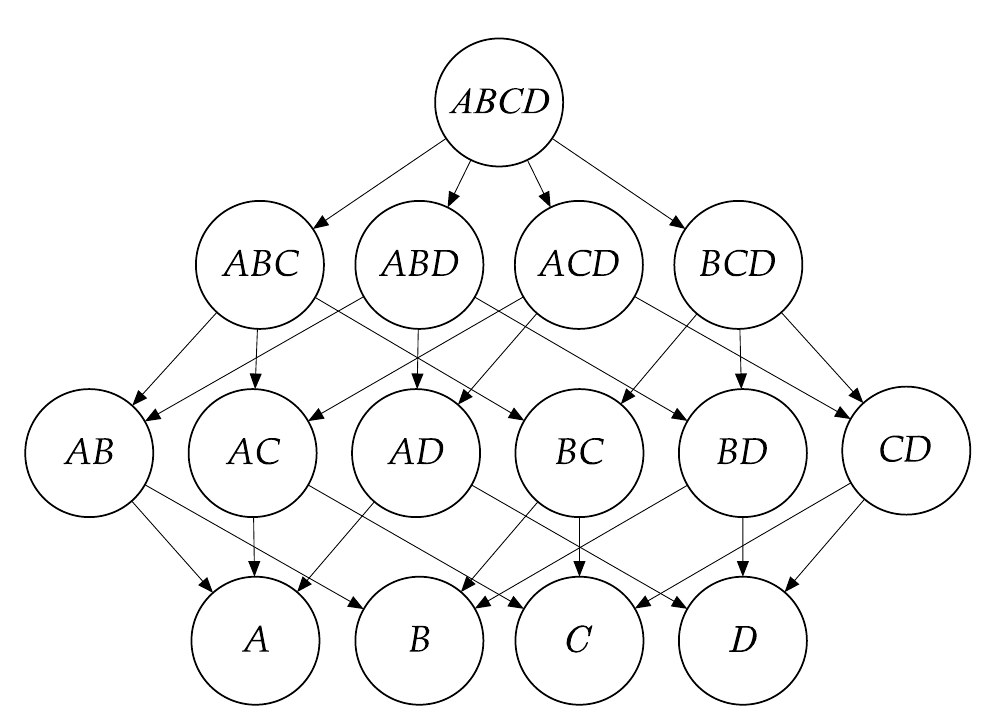}
    \caption{Belief Evolution Network under FoD $X=\{A,B,C,D\}$}
    \label{BEN_3}
  \end{minipage}
\end{figure}

Gordon and Shortliffe \cite{gordon1985method} stratify the focal elements, similar to the granular computation, by placing focal elements of the same cardinality on the same layer for reasoning. In this paper, we introduce the causality relationship in to HHS and denote it as Belief Evolution Network (BEN). Figure \ref{Bool_3} shows the process of implement mass function under $4$-element FoD on quantum circuits \cite{zhou2021quantum}. The amplitude in the quantum state reflects the 
belief of the corresponding focal element (using the binary way of encoding focal elements in \cite{smets2002application}), and we manipulate the amplitude in the quantum state by means of the RY gate. In order to make each operation not affect other focal elements, only one element is operated at a time while controlling the other elements. Hence, the evolutionary order of the focal elements is $m(ABCD)\rightarrow m(ABC)\rightarrow m(AB) \rightarrow m(A)$. In the decision process, suppose the agent has belief $a$ in $F$. When more information is obtained, $a$ only can be assigned to subsets of $F$. Based on the above, we argue that the process of belief evolved from coarse to fine can be described as the process of focal element cardinality reduction, and this process can be more strictly described as only one element reduction per evolution. They can be evolved to smaller subsets only if the coarser granularity contains the belief. Therefore, we introduce this causality into the HHS to propose a networked structure of the belief function called Belief Evolution Network (BEN).

\begin{definition}[BEN]
\rm{
For an $n$-element FoD $\Theta$, there are $n$ layers in BEN.The arrows point from the subset with larger cardinality to the subset with smaller cardinality, and the difference in cardinality between the two subsets connected by each arrow is $1$. Its root node is the FoD $\Theta$, the child nodes are the subsets whose cardinality satisfies $|F|\in (1,n)$, and the leaf nodes are singletons. For the power set in ERA and the empty set in TBM, which do not appear in BEN because there is no evolutionary relationship between them and other focal elements.
}
\end{definition}
For example, Figure \ref{BEN_3} shows the BEN under the $4$-element FoD, which clearly demonstrates the potential causality of belief in the evolutionary process.

\subsection{Full causality function}

In addition the mass function, $Bel$ functions and $Q$ functions also can be captured from BEN. For a node $F$, its $Q$ function and $Bel$ function are the sum of its parent nodes and child nodes respectively. From the perspective of BEN, both $Q(F)$ and $Bel(F)$ partially reflect the belief of focal elements which have causal relationships with $F$. Based on this, we propose a new belief function that reflects the full causal relationship of focal elements. 

\begin{definition}[$FC$ function]\label{fc}
\rm{For an $n$-element FoD $\Theta$ with a BPA $m$, the full causality ($FC$) function is defined as
\begin{equation}\label{fce1}
FC(F)=\sum_{G\subseteq F;F\subseteq G}m(G)=b(F)+q(F)-m(F).
\end{equation}
When $F$ is $\Theta$, $FC(F)$ degenerates as $b(F)$; when $F$ is empty set, $FC(F)$ degenerates to $q(F)$.}
\end{definition}

According to the Eq. (\ref{fc}), $FC$ function can be regarded as the complete set of $b$ function and $q$ function. Based on the Eq. (\ref{bel_eq}), the belief functions and mass function can realize reversible transformations. Similarly, $FC$ function also can be transformed to mass function. Based on the matrix operation \cite{smets2002application}, it can be written as $\boldsymbol{FC}=\boldsymbol{m2FC}\cdot\boldsymbol{m}$, where $\boldsymbol{m2FC}$ is $\boldsymbol{m2FC}(F,G)=\begin{cases}
1 & F\subseteq G;G\subseteq F \\
0 & \text{others}
\end{cases}$. Analyzed from the perspective of physical meaning, $FC(F)$ denotes the sum of the belief of all focal elements for which there is a causal relationship during the evolution. When credible information is received, the belief of $F$ flows to its child nodes, and when unreliable information is encountered, the belief of $F$ is discounted to its parent nodes. Compared with plausibility functions, plausibility cannot be evolved by information fusion or discounting directly. Hence, the $FC$ function more emphasizes the inclusion relationships of focal elements rather than intersection. 

\subsection{Discussion}
In this section, we introduce causality into the HHS and propose BEN to represent the potential evolutionary relationships between focal elements. Based on BEN, we propose a new belief function, called the full causal ($FC$) function, to represent the sum of belief which excites causal relationships with focal elements. Compared with the previous idea of dealing with power sets \cite{zhou2022fractal}, BEN narrows down the transfer of belief to between one element to portray a more detailed belief evolution process.

\section{Full Causality Probability Transformation}
\label{benpts}

In this section, we interpret the PT methods on BEN firstly, and propose a new PT method based on the BEN and $FC$ function, Under the Bi-Criteria evaluation, the proposed method is superior to existing PT methods.

\subsection{Interpret Probability Transformation on Belief Evolution Network} 

In \cite{zhou2020fractal}, we propose a process of PPT based on assigning belief of focal elements on their power sets continuously. From the perspective of the BEN, the PPT also can be modeled in a belief evolution view. For an $n$-element FoD $\Theta$ with a BPA $m$, after model the $m$ on BEN, starting from the top layer, the belief transfer of each node is assigned evenly to all its child nodes in the next layer, and after $n-1$ times evolution, the belief finally falls in the bottom layer (singletons), which corresponds to the unnormalized $BetP$ \footnote{For the normal BPA, it is $BetP$; and for the subnormal $BetP$, it is unnormalized PPT.}. According to the introduction, PT also can be interpreted from the perspective information fusion. For the PTM, it can be written as $PnPl=m\oplus P_\mathrm{N}$, where $P_\mathrm{N}$ is a uniform distribution $P_\mathrm{N}(\theta)=\frac{1}{n}$. However, PPT cannot be written as the form of DRC directly, and we propose a partial DRC.

\begin{definition}[partial DRC]\rm{
For two BPAs $m_1$ and $m_2$ under an $n$-element FoD $\Theta$, its partial DRC is defined as
\begin{equation}\label{pdrc}
    m_1\oplus_\mathrm{p}m_2(F)=\sum_{G\cap H =F; G\subset H; G\in m_1; H\in m_2}m_1(G)\frac{m_2(H)}{\sum_{I\subset G}m_2(I)}+m_2(F).
\end{equation}
It only fuses part of the information in the fusion process, not all of it, so it is named partial DRC. Since partial DRC does not satisfy the exchange law, i.e., $m_1\oplus_\mathrm{p}m_2\neq m_2\oplus_\mathrm{p}m_1$, it is defined in an ordered manner.}
\end{definition}

According to Eq.(\ref{pdrc}), the PPT can be written as a process of information fusion. For the original BPA $m$, after executing $n-1$ times partial DRC, its result is $BetP$. The $t$th fusion is $m=m_t\oplus_\mathrm{P}m$, where $m_t(F)=\begin{cases}
1/\binom{n}{n-t}& |F|=n-t\\0&\text{others}
\end{cases}$. For example, given a BPA $\{m(AB)=0.3,m(BC)=0.1,m(ABC)=0.6\}$, based on partial DRC, its PPT can be calculated as
$$m=m_1\oplus_\mathrm{P}m=\{m(AB)=0.5,m(AC)=0.2,m(BC)=0.3\};$$
$$m=m_2\oplus_\mathrm{P}m=\{m(A)=0.35,m(B)=0.4,m(C)=0.25\};$$
which is equal to $BetP$. Hence, the partial DRC also can simulate the belief evolution on BEN. In addition, other PT methods also can be modeled on the BEN.

\begin{figure}[htbp!]
\centering
\includegraphics[width=0.4\textwidth]{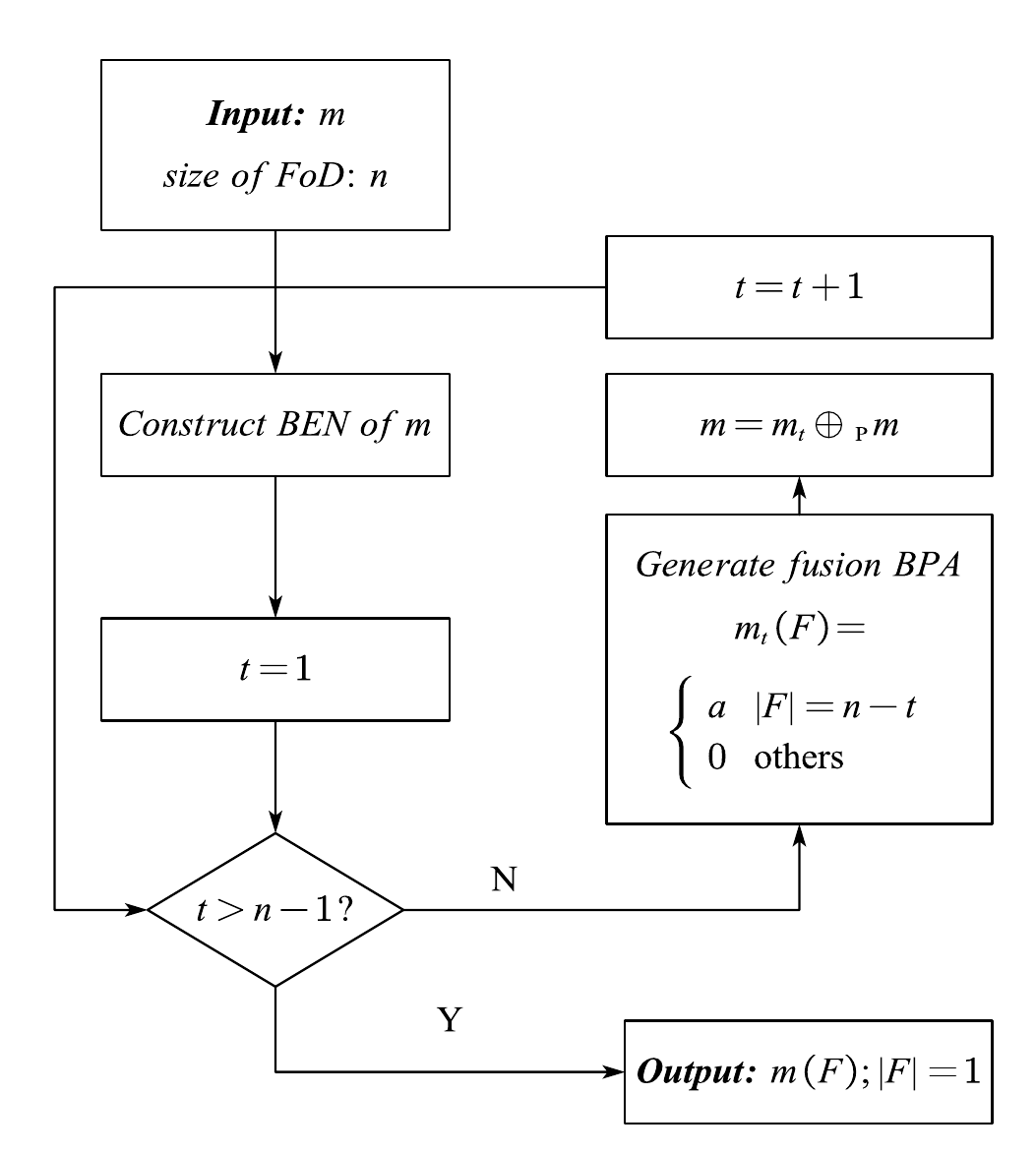}
\caption{Generalized Probability Transformation Model on BEN}
\label{ptf}
\end{figure}

According to the above description, people’s events cognition can be simulated on BEN as process of completely unknown (top layer) evolution to precise description (bottom layer). In the process of evolution, BPA can be represented on BEN, which means the cognitive state of people at this moment, but in actual observations, what people get at this moment is a probability distribution. This process from BPA to PMF can be interpreted as PT method, which utilizes the known information to predict the remaining evolution process on BEN, and transform the belief to bottom layer. Based on above, the generalized PT model (GPTM) is summarized in Figure \ref{ptf}. Different from PPT, other PT methods assign the belief of focal elements with a cardinality greater than $1$ to singletons in some proportion, so there is no way for them to be written directly in the form of partial DRC, or for them to evolve on BEN in more than one way. PT can be seen as a process of information fusion, and since there is no more information, both PPT and PTM fuse the uniform distribution to represent ignorance. For other methods, they generate the fused BPA $m_t$ by mining more information from original BPA $m$. Hence, under the BEN's interpretation, determining $m_t$ is a significant premise for proposing a reasonable PT method.

\subsection{Full Causality Probability Transformation}

Based on the interpretation of PT on BEN, we propose a new PT method based on the BEN and full causality function. We argue that the $FC$ function can reflect the potential evolutionary belief and it should be used to generate the fused BPA $m_t$. For an $n$-element FoD with its BPA $m$, we define the calculation of Full Causality Probability Transformation (FCPT) by two methods. First, based on the splitting on the BEN, the FCPT can be calculated by follow steps.

\begin{description}\label{fcp_d}
\item[\textbf{Step 1:}] Construct the $n$-element BEN and assign belief to corresponding nodes. 
\item[\textbf{Step 2:}] Iterate through BEN layer from $l=1\rightarrow (n-1)$, looping through Steps $3-5$.
\item[\textbf{Step 3:}] Calculate the $FC$ functions of subsets in $(l+1)$th layer, denoted them as $FC_{l+1}(i)$, where $i\in[1,\binom{n}{n-l}]$ means the $i_th$ subset in $(l+1)$th layer.
\item[\textbf{Step 4:}]Calculate the normalized $FC$ function of subset in $l$th layer. $FC_{(l+1),l\text{th}\_layer(k)}^\mathrm{N}(i)=\frac{FC_{l+1}(i)}{\sum_{j\subset k}FC_{l+1}(j)}$, which means the normalized $FC$ function of $k$th subset in $(l+1)$ layer under the $j$th focal element in $l$th layer.
\item[\textbf{Step 5:}] Assign the belief in $l$th layer to its child nodes according to the evolution weights $m_{l+1}(i)=m_{l+1}(i)+\sum_{i\subset k}FC_{(l+1),l\text{th}\_layer(k)}^\mathrm{N}(i) \cdot m_{l}(F_{k})$.
\item[\textbf{Step 6:}] output the belief in the $n$th layer as FCPT's result.
\end{description}

In addition, FCPT also can be defined based on the GPTM, where $m_t$ is defined based on $FC$ function $m_t(F)=\begin{cases}
FC(F)/\sum_{|G|=t+1}FC(G) & |F|=t+1\\
0 & \text{others}
\end{cases}$. Example \ref{e1} give a process of belief evolution in FCPT, which shows that how to assign the belief in FCPT is based on the $FC$ function and adjusting dynamically.

\begin{example}
\label{e1}

For a $4$-element FoD $X=\{A,B,C,D\}$, a BPA $m$ is:
\begin{enumerate}
\item \textbf{$1$-element focal element:}
$$m(A)=0.16,m(B)=0.14,m(C)=0.01,m(D)=0.02;$$
\item \textbf{$2$-element focal element:}
$$m(AB)=0.20,m(AC)=0.09,m(AD)=0.04,m(BC)=0.04,m(BD)=0.02,m(CD)=0.01;$$
\item \textbf{$3$-element focal element:}
$$m(ABC)=0.10,m(ABD)=0.03,m(ACD)=0.03,m(BCD)=0.03;$$
\item \textbf{$4$-element focal element:}
$$m(ABCD)=0.08.$$
\end{enumerate}

Based on the calculation method of FCPT, the transformed PMF $FCP$ is $$FCP(A)=0.4787,FCP(B)=0.3702,FCP(C)=0.0985,FCP(D)=0.0526,$$ and the specific evolution process on BEN is shown in Figure \ref{ex1f}.

\begin{figure}[htbp!]
\centering
\includegraphics[width=0.78\textwidth]{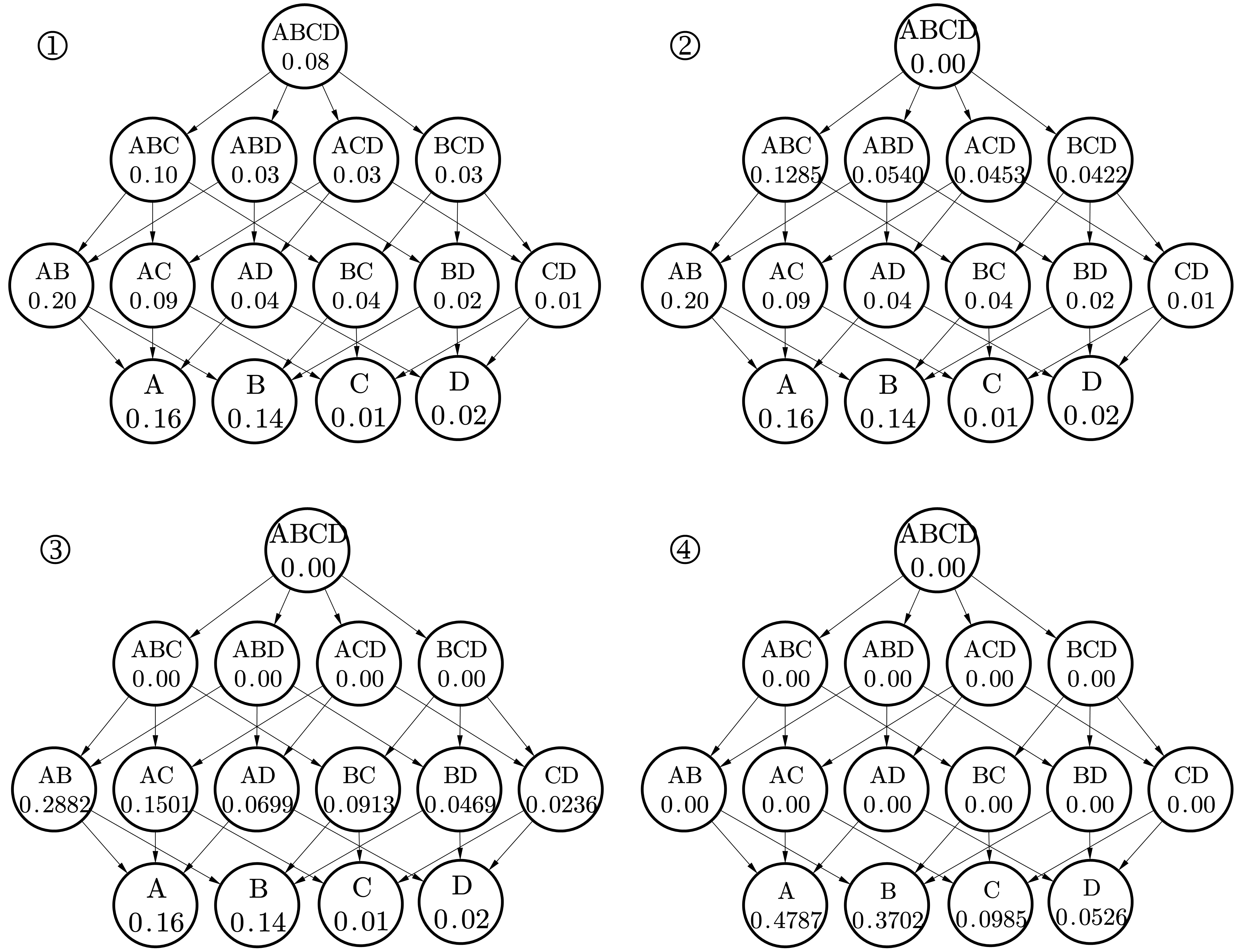}
\caption{The evolution process in Example \ref{e1}.}
\label{ex1f}
\end{figure}

\end{example}

FCPT is proposed based on BEN and $FC$ function. It determines how the focal element allocates its belief to its child nodes based on the $FC$ function. Since BEN has the structural properties of HHS, it allows the belief evolution process to dynamically adjust the proportion of belief allocation, which is the most significant difference of FCPT compared with other methods. According to Example \ref{e1}, the calculation of FCPT is more complex compared to other methods, however, this sacrifice of complexity is necessary as the final step at the decision level, if it can achieve better results. In the following part, we evaluate the performance of FCPT according to the Bi-Criteria method.

\subsection{Evaluate the FCPT}

\subsubsection{Recognition result and PIC value}

According to Eq. (\ref{pice}), when decision-making, on the premise that the recognition result is correct, a larger PIC value is more favorable for decision-making.

\begin{example}\label{pic_r_e}\rm{
For a $3$-element FoD $X=\{A,B,C\}$ with BPA $m=\{m(A)=0.1,m(AB)=0.2,m(BC)=0.3,m(ABC)=0.4\}$, the results and PIC value of transformation methods in Table \ref{s2t1-PT} $BetP$, $PnPl$, $PraPl$, $CuzzP$, $DSmP_{0.1}$, $ITP$ and $FCP$ are shown in Table \ref{pic_r_e_t}.

\begin{table}[htbp!]
\begin{center}
\begin{tabular}{c|ccc|c|c}
  \Xhline{1.4pt}
  $Methods$&$\mathbb{P}(A)$&$\mathbb{P}(B)$&$\mathbb{P}(C)$&Recognition Result&$PIC(\mathbb{P})$\\
  \hline
  $CuzzP$&$0.3455$&$\textbf{0.3681}$&$0.2864$&$\{B\}$&$0.0050$\\
  $PnPl$&$0.3043$&$\textbf{0.3913}$&$0.3043$&$\{B\}$&$0.0067$\\
  $BetP$&$0.3333$&$\textbf{0.3833}$&$0.2834$&$\{B\}$&$0.0068$\\
  $DSmP_{0.1}$&$0.3591$&$\textbf{0.3659}$&$0.2750$&$\{B\}$&$0.0073$\\
  $PraP$l&$\textbf{0.3739}$&$0.3522$&$0.2739$&$\{A\}$&$0.0077$\\
  $FCP$&$0.2951$&$\textbf{0.4688}$&$0.2361$&$\{B\}$&$0.0387$\\
  $ITP$&$\textbf{0.4140}$&$0.3885$&$0.1975$&$\{A\}$&$0.0418$\\
  \Xhline{1.4pt}
\end{tabular}
\end{center}
\caption{The transformation results of Example \ref{pic_r_e}}
\label{pic_r_e_t}
\end{table}
}
\end{example}

In Example \ref{pic_r_e}, although the PIC value of $ITP$ is the largest, its recognition result is different from most classical methods. This is because it over-utilizes the belief of single-element focal element. In the methods of the recognition result is $\{B\}$, the PIC value of $FCP$ is significantly greater than other methods, which proves its effectiveness in decision-making.

\subsubsection{Bi-Criteria evaluatione}

In previous studies on PT, PIC value is usually used as an important evaluation metric, with a higher PIC value implying less uncertainty, which is usually considered more favorable for decision making. However, a high PIC can also lead to overly optimistic decision making, making the difference between the transformed PMF and the original BPA too large. Therefore, evaluation of PT methods should consider both PIC and similarity, and since they usually have difficulty converging to the optimal value at the same time, the performance of PT can be more adequately assessed using Bi-Criteria \cite{han2015evaluation}. According to the entropy index in Eq. (\ref{ente}), using PIC value can be written as follows: 

\begin{equation}\label{piceee}
PIC'((\mathbb{P}_i))=\frac{max(PIC)-PIC(\mathbb{P}_i)}{max(PIC)-min(PIC)},
\end{equation}

so the Bi-Criteria evaluation method is 

\begin{equation}\label{hanee}
C_\mathrm{joint}((\mathbb{P}_i))=\alpha \cdot d'(\mathbb{P}_i)+(1-\alpha)\cdot PIC'(\mathbb{P}_i).
\end{equation}

We expect that the PMF after transformation has a larger PIC value (smaller $PIC'$) and a smaller distance between original BPA and transformed PMF, so a smaller $C_\mathrm{joint}$ value means the PT method. The $BetP$, $PnPl$, $CuzzP$, $DSmP$, $PraPl$ and $FCP$ of $m$ in Example \ref{e1} and their evaluation indexes are shown in Table \ref{Bi_c_t}.

\begin{table}[htbp!]

\begin{center}
\begin{tabular}{c|cccccccc}
  \Xhline{1.4pt}
  $Methods$&$\mathbb{P}(A)$&$\mathbb{P}(B)$&$\mathbb{P}(C)$&$\mathbb{P}(D)$&$PIC(\mathbb{P})$&$PIC'(\mathbb{P})$&$d(\mathbb{P})$&$d'(\mathbb{P})$\\
  \hline
  $PnPl$&$0.3614$&$0.3168$&$0.1931$&$0.1287$& $0.0526$&$1.0000$&$0.2504$&$0.1204$\\
  $CuzzP$&$0.3860$&$0.3382$&$0.1607$&$0.1151$& $0.0790$&$0.8974$&$0.2465$&$0.0087$\\
  $BetP$&$0.3983$&$0.3433$&$0.1533$&$0.1050$& $0.0926$&$0.8446$&$0.2462$&$0.0000$\\
  $PraPl$&$0.4021$&$0.3523$&$0.1394$&$0.1062$&$0.1007$&$0.8131$&$0.2464$&$0.1792$\\
  $DSmP_{0}$&$0.5176$&$0.4051$&$0.0303$&$0.0470$&$0.3100$&$0.0000$&$0.0163$&$1.0000$\\
  $DSmP_{0.001}$&$0.5162$&$0.4043$&$0.0319$&$0.0477$&$0.3058$&$0.0163$&$0.2801$&$0.9798$\\
  $FCP$&$0.4787$&$0.3702$&$0.0985$&$0.0526$&$0.2039$&$0.4122$&$0.2590$&$0.3699$\\
  \Xhline{1.4pt}
\end{tabular}
\end{center}
\caption{The results of probability transformation methods and their evaluation indexes.}
\label{Bi_c_t}
\end{table}

\begin{figure}[htbp!]
\begin{flalign}
\includegraphics[width=0.73\textwidth]{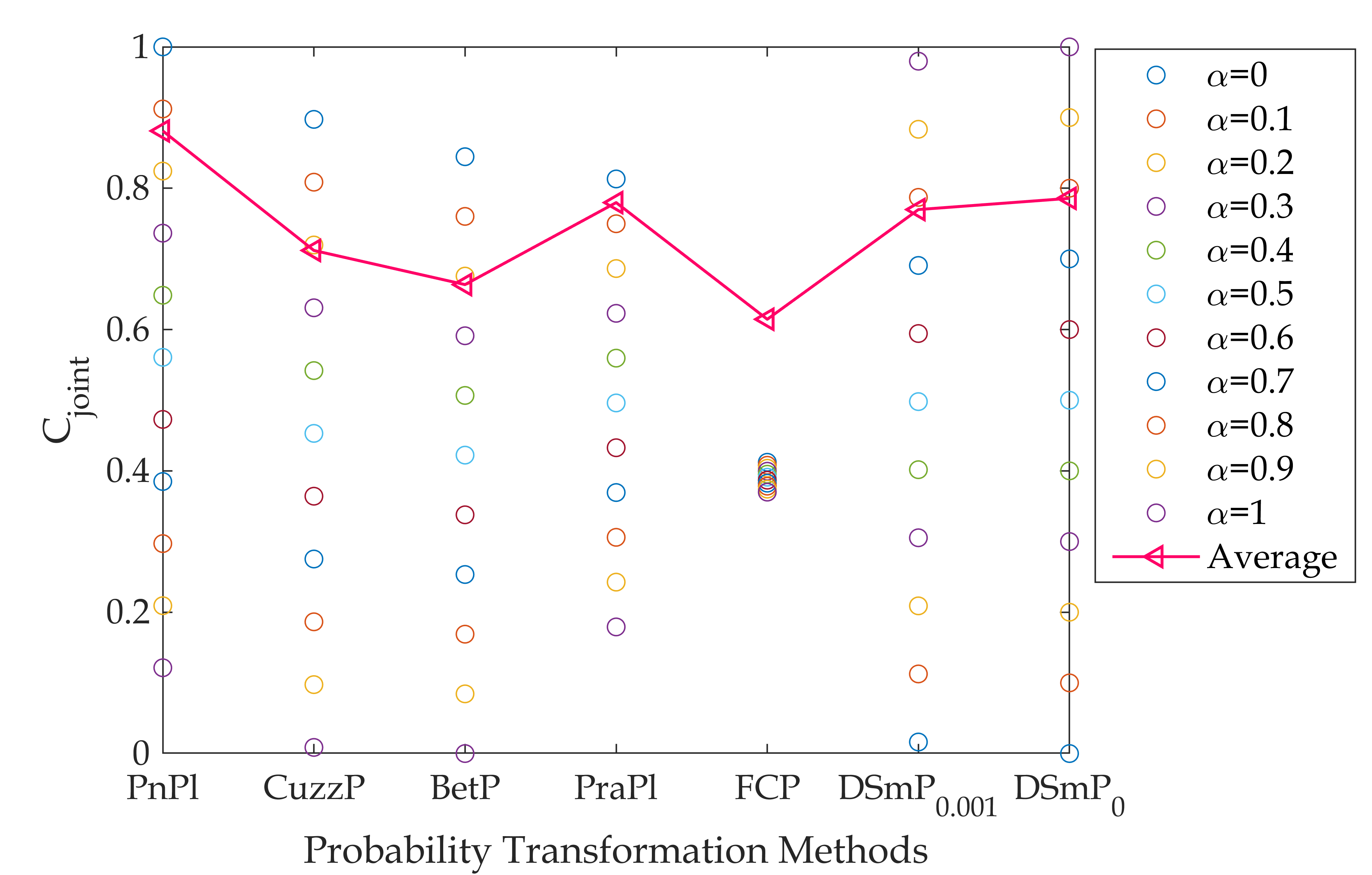}
\end{flalign}
\caption{The $C_{joint}$ of methods under different $\alpha$.}
\label{Bi_c_f}
\end{figure}

After sampling $11$ values of $\alpha$ from $0$ to $1$, the $C_{joint}$ of them are shown in the Figure \ref{Bi_c_f}. The circle represents the corresponding $C_{joint}$ under different $\alpha$, and the triangle represents the average of all the sampled values. According to Table \ref{Bi_c_t} and Figure \ref{Bi_c_f}, we can find that the other methods except FCPT have large difference between PIC index and distance index. Therefore, the total assessed value varies significantly when different values are taken for $\alpha$. FCPT takes both PIC and distance indexes into account and has the smallest result on the average value. When $\alpha=0.5$, $C_\mathrm{joint}(FCP)$ is also the smallest. Therefore, for the BPA in Example \ref{e1}, FCPT has better performance than other methods under Bi-Criteria evaluation method.

In Example \ref{r_BPA_PIC_e}, we utilize Bi-Criteria to evaluate the BPA methods of dynamic BPA qualitatively, which further demonstrates the superiority of our proposed method. The correlation coefficient \cite{jiang2018correlation} proposed by Jiang has been proven can measure similarity between BPAs reasonably, so we use it to measure the similarity between original BPA and transformed PMF.

\begin{example}\label{r_BPA_PIC_e}\rm{
For a $10$-element FoD $\Theta=\{\theta_1,\cdots,\theta_{10}\}$ with its BPA $$m: \{m(\theta_3\theta_4\theta_5)=0.15,m(\theta_6)=0.05,m(\Theta)=0.1,m(A)=0.7\},$$ when $\{A\}$ change from $\{\theta_1\}$, $\{\theta_1\theta_2\}$ to $\{\Theta\}$, the PIC values and correlation coefficients of $CuzzP$, $PnPl$, $BetP$ and $FCP$ are shown in Tables \ref{r_BPA_PIC_e_t_r} and \ref{r_BPA_PIC_e_t_PIC}, and the change trends are shown in Figures \ref{r_BPA_ex1} and \ref{PIC_ex1}.

\begin{table}[htbp!]
\begin{center}
\begin{tabular}{c|cccccccccc}
  \Xhline{1.4pt}
  $|A|$&$1$&$2$&$3$&$4$&$5$&$6$&$7$&$8$&$9$&$10$\\
  \hline
$FCP$&$0.9714$&$0.6967$&$0.5853$&$0.5127$&$0.4639$&$0.4460$&$0.4103$&$0.3768$&$0.3456$&$0.3172$\\
$BetP$&$0.9723$&$0.6976$&$0.5903$&$0.5256$&$0.4808$&$0.4625$&$0.4355$&$0.4129$&$0.3938$&$0.3772$\\
$CuzzP$&$0.9712$&$0.6807$&$0.5864$&$0.5239$&$0.4793$&$0.4608$&$0.4333$&$0.4102$&$0.3903$&$0.3731$\\
$PnPl$&$0.8920$&$0.6831$&$0.5870$&$0.5234$&$0.4783$&$0.4602$&$0.4326$&$0.4092$&$0.3891$&$0.3717$\\
  \Xhline{1.4pt}
\end{tabular}
\end{center}
\caption{The correlation coefficients in Example \ref{r_BPA_PIC_e}.}
\label{r_BPA_PIC_e_t_r}
\end{table}

\begin{table}[htbp!]
\begin{center}
\begin{tabular}{c|cccccccccc}
  \Xhline{1.4pt}
  $|A|$&$1$&$2$&$3$&$4$&$5$&$6$&$7$&$8$&$9$&$10$\\
  \hline
$FCP$&$0.6365$&$0.4031$&$0.3622$&$0.3386$&$0.2824$&$0.2689$&$0.2338$&$0.2185$&$0.2191$&$0.2305$\\
$BetP$&$0.5011$&$0.3073$&$0.2451$&$0.2066$&$0.1801$&$0.1609$&$0.1143$&$0.0753$&$0.0419$&$0.0127$\\
$CuzzP$&$0.4891$&$0.1372$&$0.1511$&$0.1467$&$0.1338$&$0.1291$&$0.0951$&$0.0631$&$0.0330$&$0.0047$\\
$PnPl$&$0.1336$&$0.1469$&$0.1657$&$0.1639$&$0.1526$&$0.1251$&$0.0913$&$0.0595$&$0.0296$&$0.0013$\\
  \Xhline{1.4pt}
\end{tabular}
\end{center}
\caption{The PIC values in Example \ref{r_BPA_PIC_e}.}
\label{r_BPA_PIC_e_t_PIC}
\end{table}

  \begin{figure}[htbp!]
  \begin{minipage}[htbp]{0.48\linewidth}
    \centering
    \includegraphics[width=0.98\textwidth]{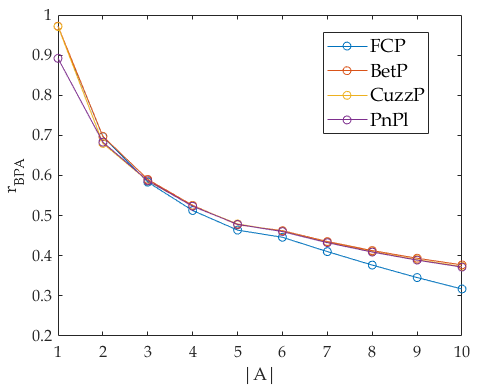}
    \caption{Trends of $r_{BPA}$ in Example \ref{r_BPA_PIC_e}.}
    \label{r_BPA_ex1}
  \end{minipage}
  \begin{minipage}[htbp]{0.48\linewidth}
    \centering
    \includegraphics[width=0.98\textwidth]{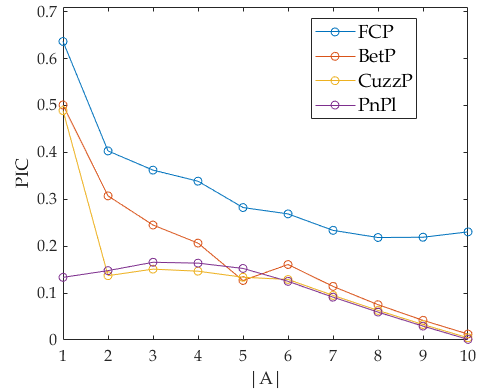}
    \caption{Trends of PIC values in Example \ref{r_BPA_PIC_e}.}
    \label{PIC_ex1}
  \end{minipage}
\end{figure}}
\end{example}

In Example \ref{r_BPA_PIC_e}, we can find that $BetP$ has the greatest correlation with original BBA, and for other methods, their correlation is very close to $BetP$. But in terms of PIC values, the results of $FCP$ is obviously greater than other methods. Although these methods are relatively close under the similarity index, $FCP$ has absolute advantages in decision-making, especially when the multi-element focal elements' belief is high.

In this section, we first give a new interpretation of PT using BEN, arguing that PT can also be seen as a process of information fusion, and propose the partial DRC to quantify this process. Second, we give a Generalized Probability Transformation Model (GPTM) based on BEN and partial DRC and use it to intpret PPT from the information fusion perspective. Based on the above interpretation, we use the $FC$ function to mine the potential tendencies in BPA and propose a new PT method called FCPT. under the evaluation method of Bi-Criteria, FCPT performs better than the other classical PT method. However, the Bi-Criteria evaluation method does not reflect the effectiveness of PT in practical applications. Therefore, we propose a new probability combination rule in the next section to show the potential application scenarios of FCPT.

\section{FCPT-PCR: Full Causality Probability Transformation-based Probability Combination Rule}
\label{DTCR}

From the previous evaluation, it is clear that FCPT can maintain a higher degree of similarity with the original BPA while mining the belief tendency, based on which we propose a new probability combination rule called Full Causality Probability Transformation-based Probability Combination Rule (FCPT-PCR). Unlike DRC, the proposed probability combination rule models the information in the framework of DST and uses FCPT to implement the fusion process.

\subsection{Full Causality Probability Transformation-based Probability Combination Rule}

Based on the discussion in Section \ref{intro}, there are two important properties for 
information combination rules should be satisfied. First, when fusing the conflict information, the proposed method should not generate counter-intuitive result. For example, there are two PMFs under a $3$-element FoD $P_1=\{0.9,0.09,0.01\}$ and $P_2=\{0.01,0.14,0.85\}$. They both give the smallest support to the second element, yet after DRC fusion the result is $P_{1\oplus 2}=\{0.2990,0.4186,0,2824\}$. Therefore, the proposed method should be in line with the intuition that, for this example, the PMF after fusion should still be that the second element has the smallest support. Second, if the same information is fused, their results should have a more pronounced tendency, however if the same information is fused consecutively, their results should not always expand the tendency. For example, if the information $P_3=\{0.5,0.25,0.25\}$ fuse itself, the result of DRC is $P_{3\oplus3}=\{0.6667,0.1667,0.1667\}$, which is intuitive. However, if we fuse $P_3$ continuously, it will be reach the $P_{3\oplus\cdots\oplus3}=\{1,0,0\}$. In practical decision making, consecutively receiving $P_3$ does not make us trust the first element more when our support for the first element is already high. Therefore, from this perspective, it is not reasonable to get probability of first element being $1$ after consecutively fusing $P_3$. Therefore, to achieve these two properties, we propose a new Probability Combination Rule based on DCR and FCPT.

\begin{definition}[FCPT-PCR]\label{dtcr_d}\rm{
For an $n$-element FoD with PMFs $P_1$ and $P_2$, the Full Causality Probability Transformation-based Probability Combination Rule (FCPT) $P_{1}\circledsmall{$\uplus$}P_{2}$ can be divided as $2$ steps.
\begin{enumerate}
\item \textbf{Step $1$:} Fuse $P_1$ and $P_2$ by Disjunctive Combination Rule (DCR): $m_{1\circledtiny{$\cup$}2}=P_{1}\circledsmall{$\cup$}P_{2}$.
\item \textbf{Step $2$:} Transform $m_{1\circledtiny{$\cup$}2}$ to $\mathbb{P}_{1\circledtiny{$\uplus$}2}$ utilizing FCPT method.
\end{enumerate}}
\end{definition}

\begin{figure}[htbp!]
\centering
\includegraphics[width=0.45\textwidth]{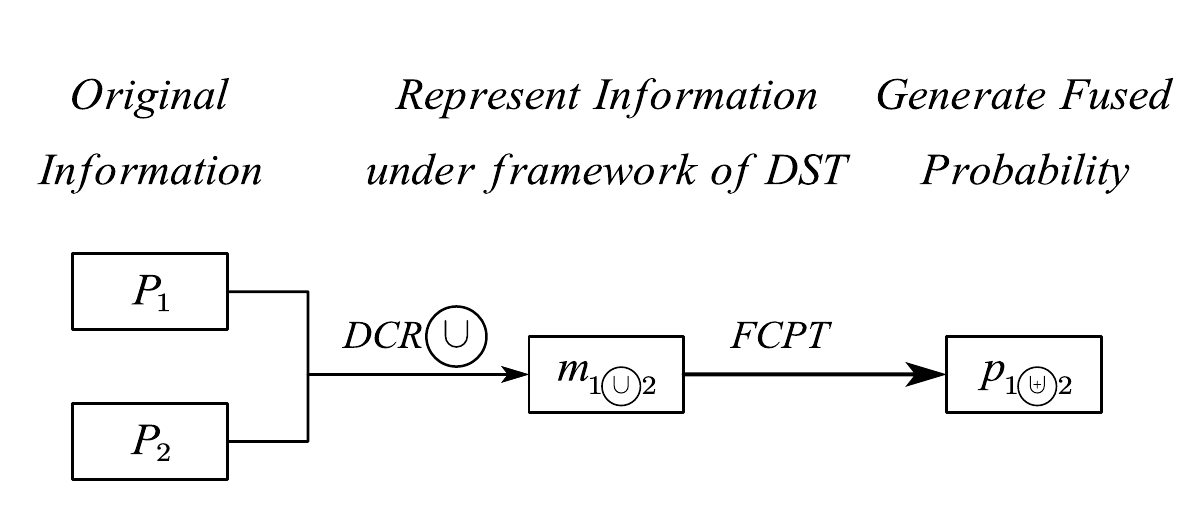}
\caption{The process of FCPT-PCR}
\label{DTCR_f}
\end{figure}

Figure \ref{DTCR_f} shows the flowchart of FCPT-PCR. Based on Definition \ref{dtcr_d}, we firstly use DCR to represent probabilistic information on the space of power set, which has better performance to resolve conflict than DRC. Secondly, we use FCPT to project the BPA to probability space, FCPT has been proven that has larger PIC value under the premise of being similar to original BPA, so it can maintain the correct belief tendency in fusion process.

DTCR is the same as CRD in terms of mathematical algorithm. It satisfies the commutative law but not the associative law. Examples \ref{conflict_e} and \ref{matai_e} are used to show its performance in two important properties.

\begin{example}\label{conflict_e}\rm{
For PMFs $P_1=\{0.9,0.09,0.01\}$ and $P_2=\{0.01,0.14,0.85\}$ under the $3$-element FoD $X=\{A,B,C\}$, using DRC and FCPT-PCR to fuse them respectively.
\begin{equation}\label{conflict_e_e}
\begin{aligned}
P_{1\oplus 2}=\{0.2990,0.4186,0,2824\};~P_{1\circledtiny{$\uplus$}2}=\{0.5046,0.0531,0.4423\}.
\end{aligned}
\end{equation}}
\end{example}

In Example \ref{conflict_e}$P_1$ and $P_2$ have high degree of support to $A$ and $C$ respectively. In the face of highly conflicting information, DRC assigns the highest support to $B$, which is obviously counter-intuitive. FCPT-PCR argues these two pieces of information indicate that $A$ and $C$ have similar support degree, and the fusion result adjust the $B$'s support degree to be lower than any piece of information. The final support degree of $A$ is higher than $C$, which satisfies the first property. Therefore, from the perspective of resolving conflict, FCPT-PCR has better performance than DRC.

\begin{example}\label{matai_e}
For PMFs $P_{1}=P_{2}=\{0.5,0.25,0.25\}$ under the $3$-element FoD $X=\{A,B,C\}$, using DCR and FCPT-PCR to fuse them respectively.
\begin{equation}\label{matai_e_e1}
\begin{aligned}
P_{1\oplus 2}=\{0.6667,0.1667,0,1667\};~P_{1\circledtiny{$\uplus$}2}=\{0.5658,0.2171,0.2171\}.
\end{aligned}
\end{equation}
We suppose that all received information is the same PMF $P=\{p(A),\frac{1-p(A)}{2},\frac{1-p(A)}{2}\}$. When the support to $A$ of the original distribution increases from $0.34$ to $1$, the $p(A)$ after $15$ times fusion by DRC and FCPT-PCR are shown in Figures \ref{matai_f} and \ref{wmatai_f}. 

\begin{figure*}[htbp!]
  \begin{minipage}[htbp]{0.48\linewidth}
    \centering
    \includegraphics[width=0.98\textwidth]{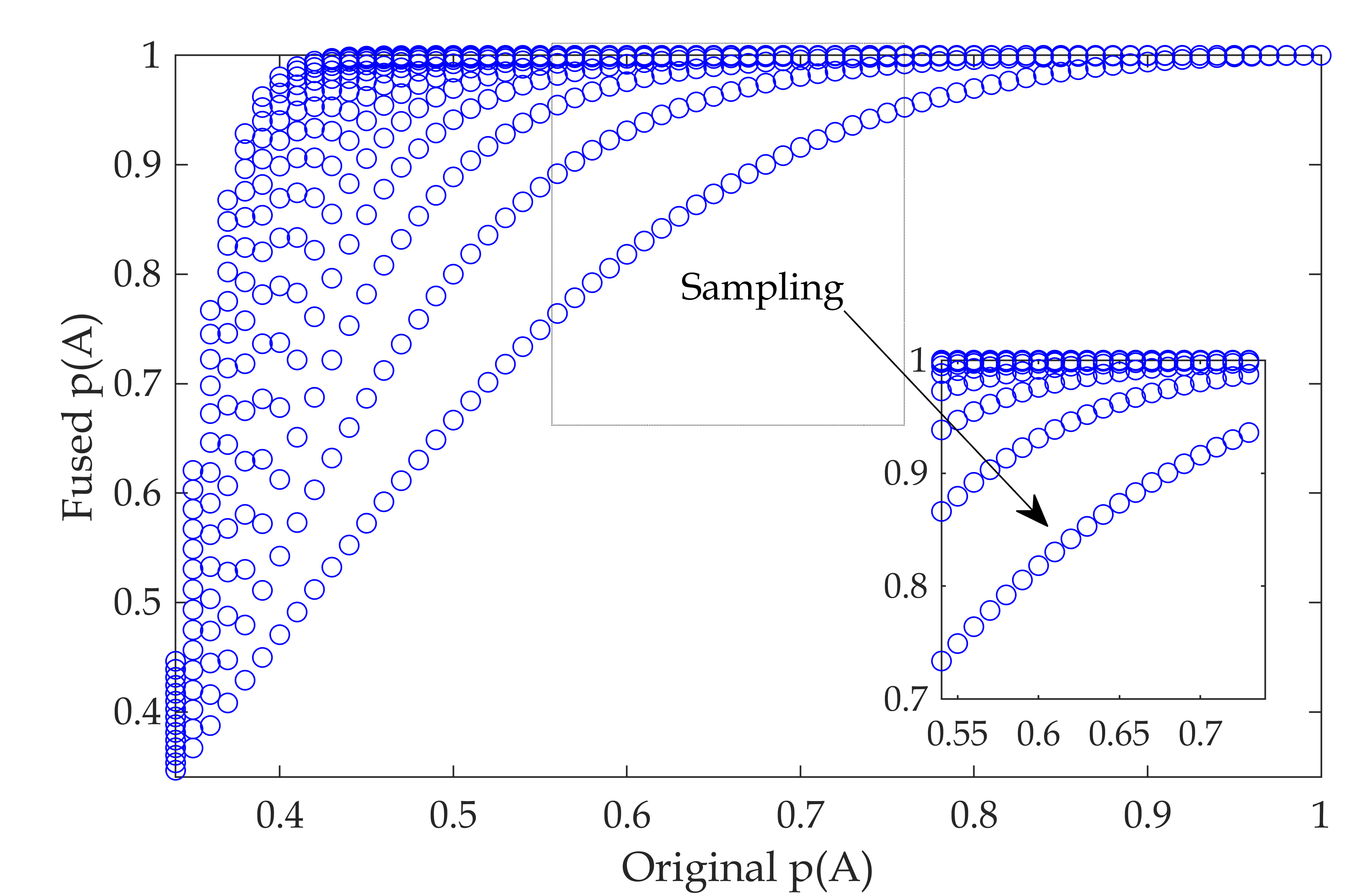}
    \caption{Fusion same probability distribution $15$ times by DRC.}
    \label{matai_f}
  \end{minipage}
  \begin{minipage}[htbp]{0.48\linewidth}
    \centering
    \includegraphics[width=0.98\textwidth]{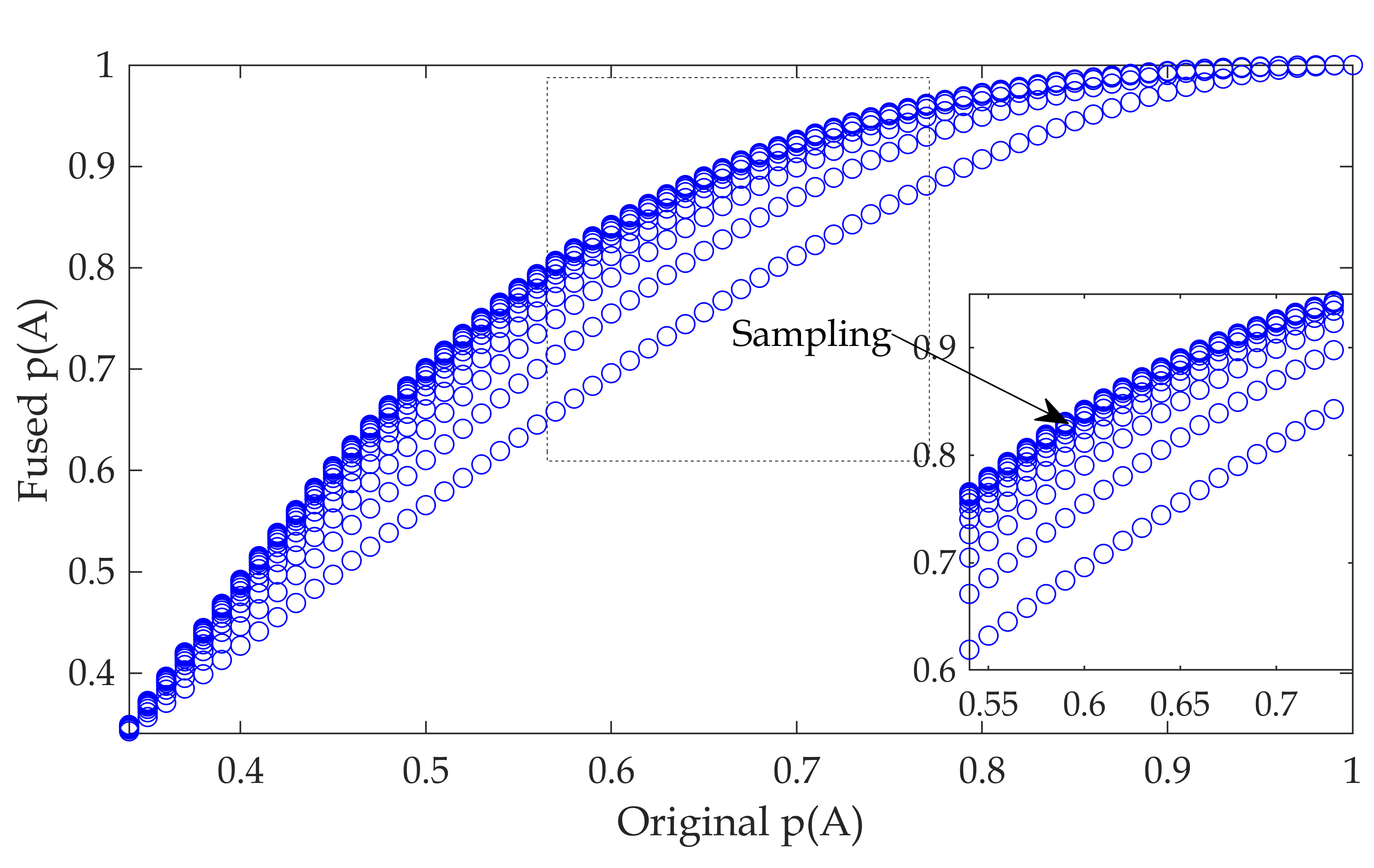}
    \caption{Fusion same probability distribution $15$ times by FCPT-PCR.}
    \label{wmatai_f}
  \end{minipage}
\end{figure*}
\end{example}

According to Equ. (\ref{matai_e_e1}), the FCPT-PCR is similar to DRC in that it amplifies tendency to object when fusing the same information, which is more convenient to make decisions. But according to Figures \ref{matai_f} and \ref{wmatai_f}, we can find that any support degree to object with tendency will reach $1$ after a limited number of DRCs, which is the Matthew effect. For FCPT-PCR, fusion the same information having object with tendency also can increase the support to the object with tendency, but it will approach a limit. We call this property the pseudo-Matthew effect. For the PMF in Example \ref{matai_e}, after limited times fusion by DRC and FCPT-PCR respectively, the $p(A)$ are $1$ and $0.7016$. In actual expert decision-making, for the three outcomes $\{A,B,C\}$, all experts believe that there is a half probability of outcome $\{A\}$. This does not allow us to determine that $\{A\}$ is the final outcome, but can magnify the probability of $\{A\}$ to more than half probability. From this perspective, the pseudo-Matthew effect is more reasonable than the Matthew effect. In summary, FCPT-PCT produces more reasonable results than DRC in terms of conflict information fusion and the same information fusion.

\subsection{Ablation experiment}

According to Figure \ref{DTCR_f}, the second step of FCPT-PCR is probability transformation. In Example \ref{xiaorongshiyan}, we replace $FCP$ by the most classic four methods: $BetP$, $PnPl$, $DSmP_0$ and $CuzzP$. The results show that only $FCP$ can realize the performance of FCPT-PCR.

\begin{example}\label{xiaorongshiyan}\rm{
We use $BetP$, $PnPl$, $DSmP_0$ and $CuzzP$ to replace $FCP$ in Definition \ref{dtcr_d}, and fusion the PMFs in Example \ref{conflict_e} and \ref{matai_e}, the results $P^1$ and $P^2$ are shown in Table \ref{xiaorongshiyan_t1}.

\begin{table}[htbp!]
\begin{center}
\begin{tabular}{c|ccccc}
  \Xhline{1.4pt}
  $Methods$&$FCP$&$DSmP_{0}$&$BetP$&$PnPl$&$CuzzP$\\
  \hline
  $P^1(A)$&$0.5046$&$0.2990$&$0.4550$&$0.4574$&$0.4550$\\
  $P^1(B)$&$0.0531$&$0.4186$&$0.1550$&$0.1104$&$0.1550$\\
  $P^1(C)$&$0.4423$&$0.2824$&$0.4300$&$0.4323$&$0.4300$\\
  \Xhline{1.4pt}
    $P^2(A)$&$0.5658$&$0.6667$&$0.5000$&$0.4574$&$0.4615$\\
  $P^2(B)$&$0.2171$&$0.1667$&$0.2500$&$0.1104$&$0.2692$\\
  $P^2(C)$&$0.2171$&$0.1667$&$0.2500$&$0.4323$&$0.2692$\\
    \Xhline{1.4pt}
\end{tabular}
\end{center}
\caption{The combination results in Example \ref{xiaorongshiyan}.}
\label{xiaorongshiyan_t1}
\end{table}}
\end{example}

In terms of resolve conflicting, $DSmP_0$ produces the similar counter-intuitive results as DRC, because it only considers the belief of singletons, so the results obtained are the same as DRC. Although the results of the other $3$ methods assign the degree of support into $A$ and $C$, in terms of the degree of support to $B$, their results are the intermediate value of the two pieces of information $(P^1(B)\in[P_1(B),P_2(B)])$, which shows that they does not reduce the probability of $B$. So all of them is unreasonable in fusing conflicting information. In terms of fusion the same information, $DSmP_0$ and DRC have the same Matthew effect. The results of $BetP$ and $CuzzP$ both are same as the results before fusion. After $PnPl$ fusion, the degree of support of $A$ is reduced. Based on above, Example \ref{xiaorongshiyan} proves sufficiently that only FCPT is adapt to compose the combination rule in Figure \ref{DTCR_f}.

\subsection{Fusion multi-source probability}

Since FCPT-PCR does not satisfy the associative law, we choose to use Murphy's idea \cite{murphy2000combining} for multi-source information fusion. For $n$ pieces of probability information, we average the distributions and use FCPT-PCR to fuse $n$ times, and the specific process is shown in Algorithm \ref{MDTCR_A}.

\begin{algorithm}[htbp]  
  \caption{Multi-source information fusion based on FCPT-PCR}  
  \label{MDTCR_A}  
  \begin{algorithmic}[1]  
    \Require
     PMFs under the $n$-element FoD $P_1\cdots P_m$;
     \Ensure
     Fused PMF $\mathbb{P}_{1\circledtiny{$\uplus$}\cdots\circledtiny{$\uplus$}2}$;  
     \For {$i=1:m-1$}
    \State $P_{i+1}=P_i+P_{i+1}$; 
    \EndFor
    \State $\overline{P}=P_m./m$; \%\textit{Get probability average.}
    \State $P_0=\overline{P}$;
    \For{i=1:m-1} \%\textit{Fusion average probability m times.}
    \State $\overline{P}=\overline{P}\circledtiny{$\uplus$}P_0$; 
    \EndFor
    \State $\mathbb{P}_{1\circledtiny{$\uplus$}\cdots\circledtiny{$\uplus$}2}=\overline{P}$;
    \State\Return $\mathbb{P}_{1\circledtiny{$\uplus$}\cdots\circledtiny{$\uplus$}2}$;  
  \end{algorithmic}  
\end{algorithm}

\begin{example}\label{MDTCR_E}\rm{
For $n$ PMFs under $4$-element FoD, we use Multi-source information fusion based on FCPT-PCR and Murphy's method to fuse them respectively. The PMFs and their fusion results are shown in Table \ref{MDTCR_E_t}.

\begin{table}[htbp!]
\begin{center}
\begin{tabular}{c|cccc|c}
  \Xhline{1.4pt}
  $Probability$&$A$&$B$&$C$&$D$&$Result$\\
  \hline
  $P_1$&$0.30$&$0.60$&$0.09$&$0.01$&$\{B\}$\\
  $P_2$&$0.30$&$0.01$&$0.01$&$0.68$&$\{D\}$\\
  $P_3$&$0.02$&$0.02$&$0.30$&$0.66$&$\{D\}$\\
  $P_4$&$0.20$&$0.10$&$0.70$&$0.00$&$\{C\}$\\
  $P_5$&$0.02$&$0.80$&$0.08$&$0.10$&$\{B\}$\\
  $P_6$&$0.60$&$0.30$&$0.05$&$0.05$&$\{A\}$\\
  $P_7$&$0.90$&$0.50$&$0.50$&$0.35$&$\{A\}$\\
  $P_8$&$0.30$&$0.30$&$0.40$&$0.00$&$\{C\}$\\
 \Xhline{1.4pt}
  Murphy's Method&$0.7685$&$0.1661$&$0.0207$&$0.0447$&$\{A\}$\\
  \hline
  Proposed Method&$0.4110$&$0.2659$&$0.1430$&$0.1802$&$\{A\}$\\
  \Xhline{1.4pt}
\end{tabular}
\end{center}
\caption{PMFs and fusion results in Example \ref{MDTCR_E}}
\label{MDTCR_E_t}
\end{table}}
\end{example}

Example \ref{MDTCR_E} shows the difference between Murphy's DRC-based method and our FCPT-PCR-based method when fusing multi-source conflicting information. Since our method and Murphy's method both use the first averaging and then the fusion method, the weight of each piece of information is equal, so we only need to compare with Murphy method to prove our methods' advantages. According to $8$ PMFs, the number of recognition results of $4$ objects is same. Intuitively, the result of fusion should have a similar degree of support to $4$ objects. From the results of the fusion, although the two methods recognize the same object $A$, our method's variance is significantly smaller than that of Murphy's. This is because the pseudo-Matthew effect is more in line with the actual human decision-making process than the traditional DRC.

\subsection{Application in Classification}

In the previous part, we discuss the difference between the proposed method and DRC by numerical examples. The rationality of FCPT-PCR in probability information fusion is illustrated in terms of two aspects: fusing conflicting information and fusing identical information. To further demonstrate its superiority, we use a classification method on probabilistic information fusion based to test the effectiveness of FCPT-PCR in practical applications. For an $n$-label classification problem, denote the label set as $\mathcal{C}=\{C_1\cdots C_n\}$. For a sample vector $\mathcal{Y}^d$ with the label $C_d$, it is jointly determined by $t$ features value $\{y^d_1,\cdots,y^d_t\}$. Its probability fusion-based method is shown as follows

\begin{enumerate}
    \item [1.] \textbf{Generate Gaussian Probability Density Function (PDF)\cite{gao2022bim}:} For each feature $f_j$, calculating its means $\mu$ and variances $\sigma^2$ of train set, and generating a Gaussian PDF $f_j(X):X\sim\mathcal{N}(\mu,\sigma)$.
    \item[2.] \textbf{Determine the PMF:} For a test sample $i$, its feature value is $x_i$. For a feature $f_j$, calculate $f_j(x_i)$ of all labels, and normalize them to determine a PMF $P_{i,j}$, which means the $j$th feature of $i$th sample. 
    \item[3.] \textbf{Information fusion:} Fusing the all PMFs of features and generate a PMF represent the probabilities of labels.
    \item[4.] \textbf{classification:} Choose the label with highest probability as the classification result.
\end{enumerate}

\begin{figure}[htbp!]
\centering
\includegraphics[width=0.8\textwidth]{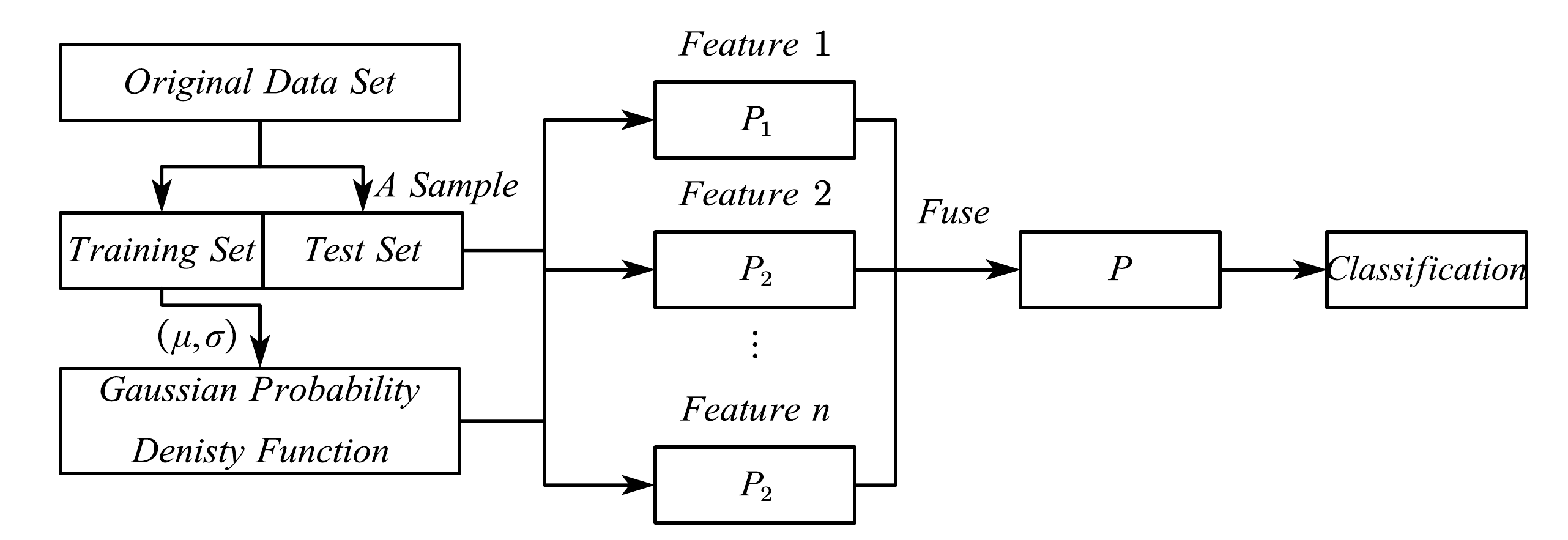}
\caption{The process of FCPT-PCR}
\label{Application}
\end{figure}

The flowchart of the method is shown in Figure \ref{Application}, and we use different probability information fusion methods (DRC, FCPT-PCR, Murphy's method) for experiments on real data sets (iris and seed, which are publicly available on the UCI \footnote{https://archive.ics.uci.edu/ml/datasets.php} dataset) to verify the effectiveness of the proposed method. To ensure the accuracy of the experimental results, we used $1000$ times $2,5,10$ fold cross-validation for each method for each data set, respectively. The specific experimental results are shown in Table \ref{app_t}, which shows that FCPT-PCR has better performance in classification than other methods.

\begin{table}[htbp!]
    \centering
    \begin{tabular}{c|ccc|c|ccc}
    \Xhline{1.4pt}
         iris&$2$-fold&$5$-fold&$10$-fold&seed&$2$-fold&$5$-fold&$10$-fold  \\
         \hline
         Murphy's method& $95.13\%$& $95.27\%$& $95.30\%$&Murphy's method& $88.84\%$& $89.02\%$& $89.06\%$\\
         \hline
         DRC& $95.21\%$& $95.33\%$& $95.33\%$&DRC& $90.37\%$& $90.38\%$& $90.33\%$\\
         \hline
         FCPT-PCR& $\textbf{95.74\%}$& $\textbf{95.83\%}$& $\textbf{95.89\%}$&FCPT-PCR& $\textbf{92.79\%}$& $\textbf{93.09\%}$& $\textbf{93.20\%}$\\
         \Xhline{1.4pt}
    \end{tabular}
    \caption{Accuracy of probability information fusion-based classification}
    \label{app_t}
\end{table}

\section{Conclusion}
\label{con}

In this paper, we introduce causality into HHS and propose a new expression of the belief function called the Belief Evolution Network. A new belief function based on the structure of the BEN is proposed, called the full causality ($FC$) function, who can represent the focal element potential transferable belief. A new general model of PT method based on BEN is proposed, and combined with the $F$C function, we propose a new PT method called Full Causality Probability Transformation (FCPT). after Bi-Criteria evaluation, FCPT can simultaneously combine higher PIC and more similar transformation results. In addition, based on FCPT, we propose a new probability fusion method and discuss the difference between it and DRC from two perspectives of fusing conflicting information and fusing identical information. In summary, the contributions of the paper are: (1) Causality is introduced in HHS to propose a BEN. (2) A new PT method based on BEN is proposed, which is superior to existing PT methods under the Bi-Criteria evolution. (3) A new probability combination rule is proposed, which has better performance than DRC in fusing conflicting information and identical information.

Apart from that, there are still some unresolved issues in this paper. Although dynamically mining the belief tendency information of BPA can yield more reasonable PT results, the computational complexity of exponential explosion will limit the FCPT for frameworks with a large number of elements. Therefore, we plan to propose an FCPT method that can perform fast evolution in our future work. Although the proposed FCPT-PCR can fuse probabilistic information better, the proposed method cannot solve the problem of DRC well for belief information. Therefore, it is also our next research goal to try to characterize the belief information in a higher dimensional space and achieve the fusion of belief information based on the idea of FCPT-PCR.

\section*{Declaration of interests}
The authors declare that they have no known competing financial interests or personal relationships that could have appeared to influence the work reported in this paper.

\section*{Acknowledgment}

The work is partially supported by National Natural Science Foundation of China (Grant No. 61973332), JSPS Invitational Fellowships for Research in Japan (Short-term).

\bibliography{myreference}

\end{document}